\documentclass[lettersize,journal]{IEEEtran}
\usepackage{booktabs}
\usepackage{multirow}
\usepackage{amsmath,amsfonts}
\usepackage{algorithmic}
\usepackage{algorithm}
\usepackage{array}
\usepackage[caption=false,font=normalsize,labelfont=sf,textfont=sf]{subfig}
\usepackage{textcomp}
\usepackage{stfloats}
\usepackage{url}
\usepackage{verbatim}
\usepackage{graphicx}
\usepackage{cite}
\begin{document}

\title{RailVQA: A Benchmark and Framework for Efficient Interpretable Visual Cognition in Automatic Train Operation}

\author{Sen~Zhang, Runmei~Li, Shizhuang~Deng, Zhichao~Zheng, Yuhe~Zhang, Jiani~Li, Kailun~Zhang, Tao~Zhang, Wenjun~Wu, and~Qunbo~Wang*
\thanks{This work was supported by the Beijing Natural Science Foundation under Grant L252146. \textit{(Corresponding author: Qunbo Wang)}}
\thanks{Sen Zhang, Runmei Li, Zhichao Zheng, Yuhe Zhang, Jiani Li, Kailun Zhang, and Qunbo Wang are with the School of Automation and Intelligence, Beijing Jiaotong University, Beijing 100091, China (e-mail: szhang02004@outlook.com; rmli@bjtu.edu.cn; zczheng7901@163.com; 24201025@bjtu.edu.cn; 24332036@bjtu.edu.cn; 24201023@bjtu.edu.cn; wangqb6@outlook.com).}
\thanks{Shizhuang Deng and Tao Zhang are with the School of Software, Northwestern Polytechnical University, Xi'an 710072, China (e-mail: dengshizhuang@mail.nwpu.edu.cn; tao\_zhang@nwpu.edu.cn).}
\thanks{Wenjun Wu is with the School of Artificial Intelligence, Beihang University, Beijing 100191, China (e-mail: wwj09315@buaa.edu.cn).}

}




\maketitle

\begin{abstract}

As Automatic Train Operation (ATO) advances toward GoA4 and beyond, it increasingly depends on efficient, reliable cab-view visual perception and decision-oriented inference to ensure safe operation in complex and dynamic railway environments. However, existing approaches focus primarily on basic perception and often generalize poorly to rare yet safety-critical corner cases. They also lack the high-level reasoning and planning capabilities required for operational decision-making. Although recent Large Multi-modal Models (LMMs) show strong generalization and cognitive capabilities, their use in safety-critical ATO is hindered by high computational cost and hallucination risk. Meanwhile, reliable domain-specific benchmarks for systematically evaluating cognitive capabilities are still lacking. To address these gaps, we introduce RailVQA-bench, the first VQA benchmark for cab-view visual cognition in ATO, comprising 20,000 single-frame and 1,168 video based QA pairs to evaluate cognitive generalization and interpretability in both static and dynamic scenarios. Furthermore, we propose RailVQA-CoM, a collaborative large–small model framework that combines small-model efficiency with large-model cognition via a transparent three-module architecture and adaptive temporal sampling, improving perceptual generalization and enabling more efficient reasoning and planning. Experiments demonstrate that the proposed approach substantially improves performance, enhances interpretability, improves efficiency, and strengthens cross-domain generalization in autonomous driving systems. Code and datasets will be available at https://cybereye-bjtu.github.io/RailVQA.html.

\end{abstract}


\section{Introduction}
\IEEEPARstart{D}{espite} increasingly standardized railway operating procedures, serious safety incidents—such as collisions and derailments caused by track intrusions, signal misinterpretation, or switch misalignment—remain challenging to prevent entirely~\cite{hong2023railway}. One contributing factor is the sustained high attentional workload placed on human drivers, which increases the likelihood of perception errors and delayed responses. Against this backdrop, Automatic Train Operation (ATO) has become essential for advancing automation and promoting the sustainable development of modern railway systems~\cite{khalil2024advanced}. Meanwhile, rapid advances in artificial intelligence have enabled the deployment of a range of driving-assistance functionalities. Yet the safety-critical nature of railway operations demands exceptional reliability, interpretability, and rigorous pre-deployment validation. This underscores the need for more intelligent train driving-assistance systems, together with a benchmark for their systematic evaluation.

Previous railway studies, such as RailSem19~\cite{zendel2019railsem19}, OSDaR23~\cite{tagiew2023osdar23}, and RailGoerl24~\cite{tagiew2025railgoerl24}, largely target low-level perception tasks (e.g., semantic segmentation, object detection, et.al). Although effective in controlled experimental settings, they are often task-specific and lack of high-level cognition, therefore generalize poorly to diverse and complex real-world scenarios. Serious safety incidents are often caused by rare long-tail anomalies rather than routine conditions, motivating a shift from basic perception to higher-level, open-world hazard cognition with robust generalization.

Recent advances in Large Multi-modal Models (LMMs)~\cite{liang2024survey, corbiere2025retrieval, alayrac2022flamingo} offer a promising pathway to address these limitations. LMMs exhibit strong open-domain visual understanding and high-level reasoning capabilities, enabling autonomous systems to interpret dynamic environments, anticipate potential hazards, and perform rule-grounded scene understanding. Furthermore, techniques such as Chain-of-Thought (CoT) reasoning~\cite{NEURIPS2022_9d560961} enhance inference reliability by explicitly structuring intermediate reasoning steps, thereby improving cognitive transparency.

Nevertheless, the direct deployment of LMMs in ATO presents substantial practical challenges. First, the inherent black-box nature of such models often results in hallucinated reasoning~\cite{liu2024survey, li2023evaluating}. Moreover, processing dense, continuous video streams with LMMs can incur relatively high computational overhead, which may make it challenging to satisfy the time-sensitivity needs of safety-critical railway operations, particularly under high-speed conditions.

More importantly, the absence of reliable and domain-specific evaluation benchmarks makes it difficult to rigorously assess the capability and safety of these models.
Due to physical constraints like fixed tracks and extended braking distances, safe train operation demands recognizing static information (e.g., signals) and anticipating dynamic risks (e.g., animals crossings or moving obstacles). Yet, current research predominantly focuses on static, single-frame inference, overlooking the dynamic, multi-frame reasoning required for reliable automatic operation ability that is more critical and better aligned with real-world ATO environments.

\begin{figure*}[htbp]
    \centering
    \includegraphics[width=\textwidth]{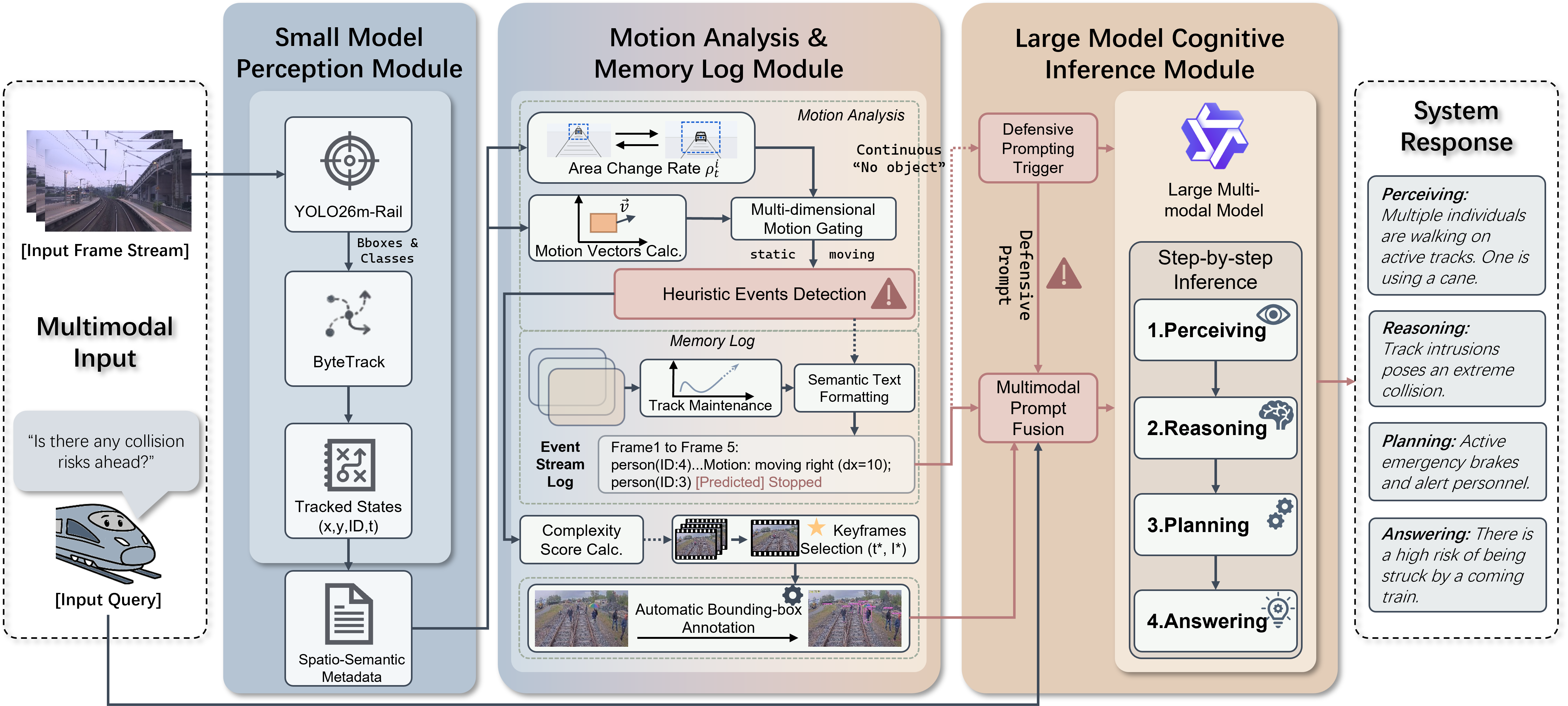}
    \caption{Overview of the proposed RailVQA-CoM framework, which consists of three hierarchical modules: (1) a Perception module that efficiently extracts visual primitives; (2) a Motion Analysis and Memory Log module that captures motion patterns and maintains temporal context for dynamic scenes; and (3) a Cognitive Inference module that performs high-level reasoning and supports decision-oriented inference.
}
    \label{fig:corail_concept}
\end{figure*}

To address these limitations, we introduce \textbf{RailVQA-bench}, the first comprehensive Visual Question Answering (VQA) benchmark specifically designed for cab-view visual cognition in ATO. RailVQA-bench establishes a unified evaluation paradigm that integrates static scene understanding with dynamic safety-critical reasoning. It comprises 20,000 single-frame QA pairs for visual information understanding and 1,168 video-based QA pairs for risk anticipation. The benchmark further requires the model to output structured perceiving–reasoning–planning Chain-of-Thought (CoT) grounded in visual evidence and operational constraints, enabling rigorous evaluation of interpretable perception and reasoning, while improving decision transparency.

We further propose \textbf{RailVQA-CoM}, a \textbf{Co}llaborative large–small \textbf{M}odel framework for cab-view visual cognition in ATO. 
As illustrated in Fig.~\ref{fig:corail_concept}, it integrates the efficiency of lightweight perception models with the cognitive capabilities of large multi-modal models (LMMs). Instead of end-to-end black-box inference, RailVQA-CoM adopts a three-module hierarchical architecture: (1) a \emph{Small-Model Perception Module} for efficient extraction of visual primitives, (2) a \emph{Motion Analysis and Memory Log Module} for analyzing motion trend and constructing semantic representations when coping with dynamic scenes, and (3) a \emph{Large Model Cognitive Inference Module}, which guides LMMs to generate structured CoT reasoning.
This complete design partly reduces computational overhead while mitigating hallucination.
Experiments demonstrate that our framework not only nearly increases end-to-end throughput speed, but also effectively mitigates fatal visual hallucinations (with up to a 10-point improvement on the evaluation metrics), while supporting model-agnostic deployment methods for LMMs.
In summary, this work makes the following three primary contributions:
\begin{itemize}
\item \textbf{Dataset:} We present the first VQA benchmark for ATO that evaluates cab-view visual cognitive generalization and interpretability. It comprises 20,000 single-image QA pairs and 1,168 video-based QA pairs.
\item \textbf{Metrics:} Building on prevailing evaluation practices in autonomous driving, we propose a 12-dimensional, logic-driven evaluation protocol with an LMM-assisted scoring pipeline for  cab-view visual cognition in ATO, shifting the focus from perception-only accuracy to cognitive reasoning.
\item \textbf{Framework:}  We develop a collaborative large–small model framework for cab-view visual cognition in ATO, enabling more efficient and interpretable decision-oriented understanding.
\end{itemize}

\section{Related Work}

\subsection{Railway Scene Perception}

Computer vision has been increasingly adopted in railway systems for basic perception tasks. To drive this field forward, the community has established several benchmarks. Early efforts, such as RailSem19~\cite{zendel2019railsem19}, partly catalyzed scene understanding in rail environments, while OSDaR23~\cite{tagiew2023osdar23} pushed the boundaries of multi-sensor fusion for obstacle detection. To further enhance perception robustness, datasets like MRSI~\cite{chen2022mrsi} introduced multi-modal infrared and RGB imagery for adverse lighting conditions, while other works have incorporated LiDAR~\cite{zhangyu2021camera} to provide precise 3D geometric measurements. More recently, RailGoerl24~\cite{tagiew2025railgoerl24} provided high-quality sequences for dynamic multi-object tracking, and SynRailObs~\cite{guo2025synrailobs} introduced synthetic data generation pipelines to enrich the distribution of rare track anomalies. However, these works predominantly target low-level perception tasks, failing to capture the causal relationship between visual elements and railway operational response.

In parallel with dataset development, algorithmic innovations are still developing. Early research frequently employed image differencing~\cite{salmane2015video} or optical flow analysis~\cite{vsilar2013obstacle} for track obstacle detection, though these methods often suffered from high false alarm rates under environmental noise. To mitigate these false alarms, modern approaches have evolved to incorporate dynamic background modeling~\cite{cao2022effective} and advanced neural networks (CNNs). For instance, the recently proposed YOLO-Rail framework~\cite{wang2026yolo} optimized real-time track obstacle detection specifically for resource-constrained edge deployment, and similar YOLO-based mechanisms have also attempted to integrate preliminary risk assessment modules~\cite{zhang2024railway}. Recent advances have introduced real-time multitask learning~\cite{chen2025real} and Lidar-Camera fusion~\cite{liu2025railfusion} to railway perception. However, the inherent rigidity of these task-customized architectures still suffer from limited generalization across open-world scenarios. Meanwhile, they often fail to couple visual features with train-specific kinematic limits, thereby falling short in complex spatiotemporal safety reasoning.

\subsection{VQA in Autonomous Driving}

Visual Question Answering (VQA)~\cite{antol2015vqa} has progressively evolved from a perception-centric task into a representative paradigm for evaluating high-level visual Cognition. This transition is largely driven by the rapid development of Large Multi-modal Models (LMMs), such as LLaVA~\cite{liu2023visual} and Qwen-VL~\cite{bai2023qwen}, which substantially enhance both visual understanding and reasoning capabilities. In contrast to traditional perception pipelines~\cite{schumann2021radarscenes}, VQA frames scene understanding and decision-oriented inference as structured question–answering, providing a more unified interface for probing and assessing complex visual reasoning processes.

Motivated by this formulation, the autonomous-driving community has adopted VQA as a benchmark and a training paradigm for vision--language reasoning in traffic scenes. Early efforts, including TrafficQA~\cite{xu2021sutd}, NuScenes-QA~\cite{qian2024nuscenes}, and LingoQA~\cite{marcu2024lingoqa}, provide structured supervision for traffic understanding. Building on this foundation, recent work integrates LMMs into end-to-end frameworks~\cite{tian2025large}. For example, DriveLM~\cite{sima2024drivelm} and DriveVLM~\cite{tian2024drivevlm} combine graph-based representations with vision–language reasoning to connect perception and planning, while DriveGPT4~\cite{xu2024drivegpt4} and RoboTron-Drive~\cite{huang2025robotron} further emphasize interpretable control and spatio-temporal alignment.
To enhance reasoning consistency, recent approaches incorporate retrieval-augmented Chain-of-Thought (CoT)~\cite{corbiere2025retrieval} and spatially-aware prompting~\cite{wu2025enhancing}, alongside more rigorous benchmarks such as DriveLMM-o1~\cite{ishaq2025drivelmm}, which evaluate step-by-step reasoning in complex driving scenarios.

Nevertheless, road-centric VQA formulations do not transfer directly to automatic train operation. Trains operate on fixed tracks, exhibit less flexibility, and must comply with stringent signaling and operating rules; these characteristics mandate customized modeling paradigms and dedicated evaluation benchmarks.

\subsection{Collaboration of Large and Small Models}

Although Large Multi-modal Models (LMMs) provide strong open-world analysis~\cite{gao2026foundation} and reasoning capabilities, the huge computational cost brought about by LMMs inference cannot be ignored. To address this issue, recent research has increasingly explored collaborative paradigms that couple large models with lightweight specialized models~\cite{chen2025survey, li2025collaborative}, enabling a more efficient allocation of computation.

A representative direction is dynamic resource allocation, as exemplified by the Big.Little Vision Transformer~\cite{guo2024big}, which improves efficiency by assigning tasks of varying complexity to models of different scales. Building on this idea, subsequent works have proposed diverse collaboration strategies. For instance, VisionGPT~\cite{kelly2024visiongpt} offloads perception to specialized modules while reserving high-level reasoning for large models, and speculative execution methods~\cite{liu2025small} accelerate inference by generating intermediate reasoning drafts with smaller models. Meanwhile, adaptive routing mechanisms, such as KCM~\cite{dai2025kcm} and AdaCoMed~\cite{chen2025multi}, dynamically assign inputs based on task difficulty, further improving efficiency and robustness. In the context of VQA, recent approaches also incorporate multi-modal heuristics to ground and stabilize LMM reasoning~\cite{sun2025large}.

Although existing large-small model collaborative architectures effectively balance efficiency and capability, they predominantly target static tasks. Currently, there is still a distinct lack of collaborative paradigms explicitly designed to improve the computational efficiency of complex reasoning on dense video streams. Achieving this is particularly difficult, as it demands a delicate trade-off between processing high-frame-rate visual inputs efficiently and maintaining the cognitive depth.

\section{RailVQA-bench}

This section details \textbf{RailVQA-bench}, the \textit{first} VQA benchmark for cab-view visual cognition in ATO. The benchmark is constructed via a pipeline combining automated generation (Sections~\ref{sec:generation}), and is compared with existing railway datasets in Table~\ref{tab:dataset_comparison}.

\subsection{Task Formulation}
\begin{figure*}[!t]
    \centering
    \includegraphics[width=0.95\textwidth]{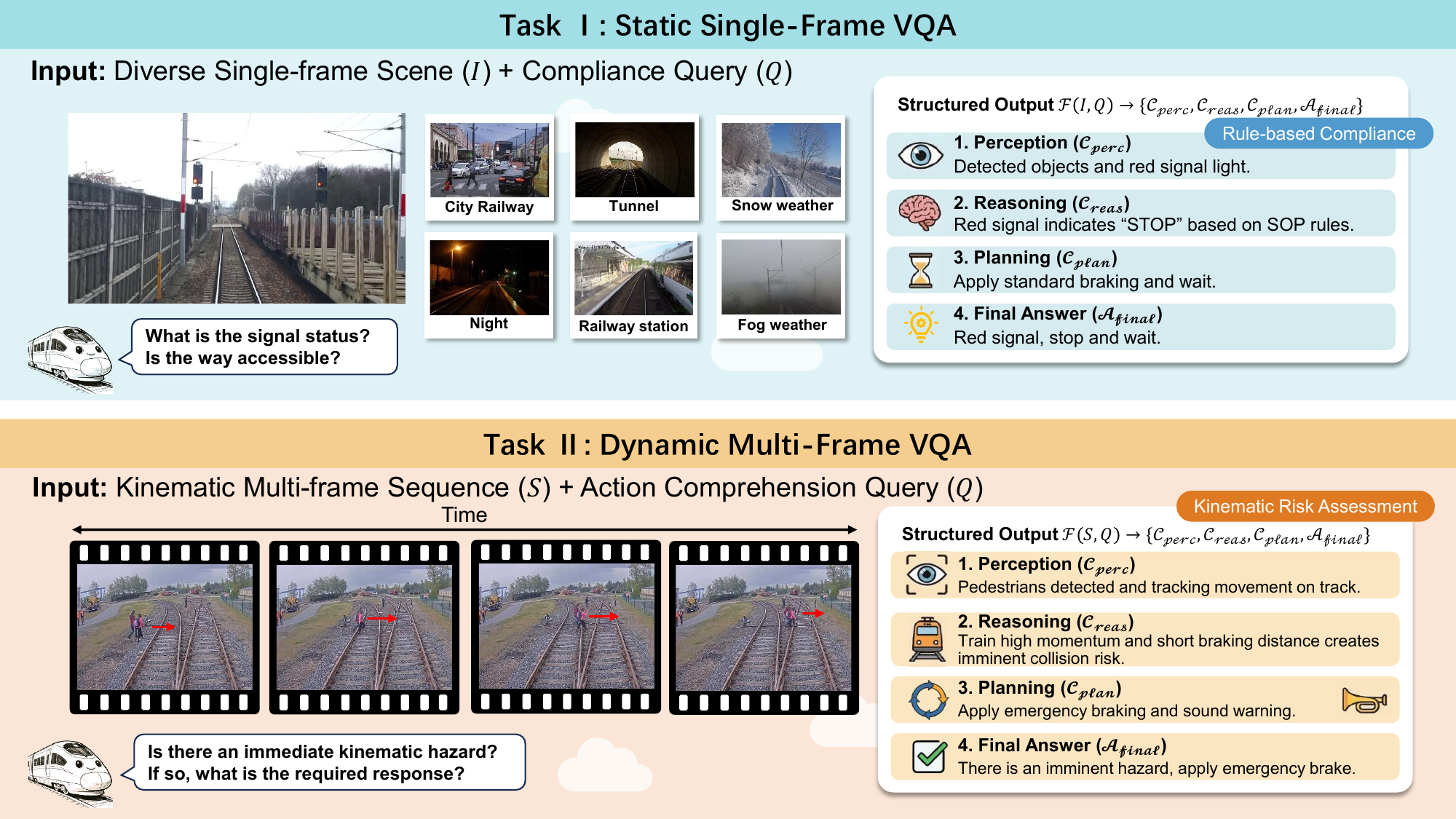} 
    \caption{This figure presents the standardized input–output schemas for the benchmark’s two core subtasks: Static Single-frame VQA and Dynamic Multi-frame VQA. Given a visual input—either a single frame $I$ or a video sequence $S$—and its associated question $Q$, the model output is required to follow a predefined, structured chain-of-thought (CoT) format.
    }
    \label{fig:railvqa_main}
\end{figure*}

\begin{figure*}[!t]
    \centering
    \captionsetup[subfloat]{font=scriptsize}
    \subfloat[Distribution of CoT Lengths]{
        \includegraphics[width=0.28\linewidth]{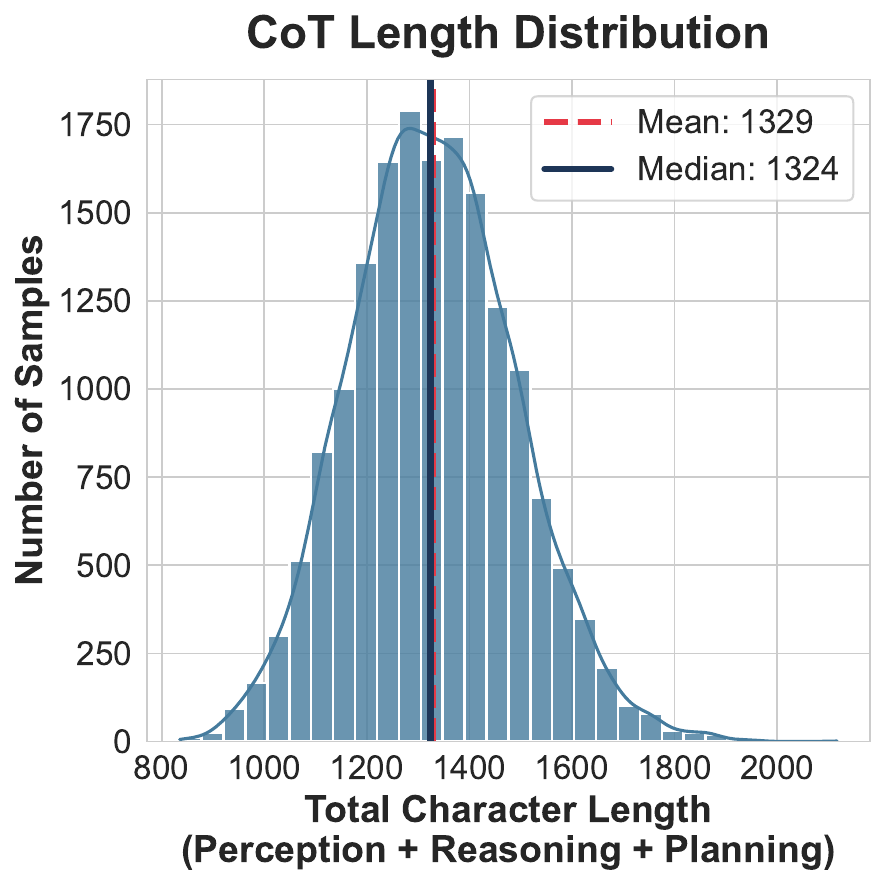}
        \label{fig:railvqa_hist}
    }
    \hfill
    \subfloat[Distribution of Key Railway Entities]{
        \includegraphics[width=0.38\linewidth]{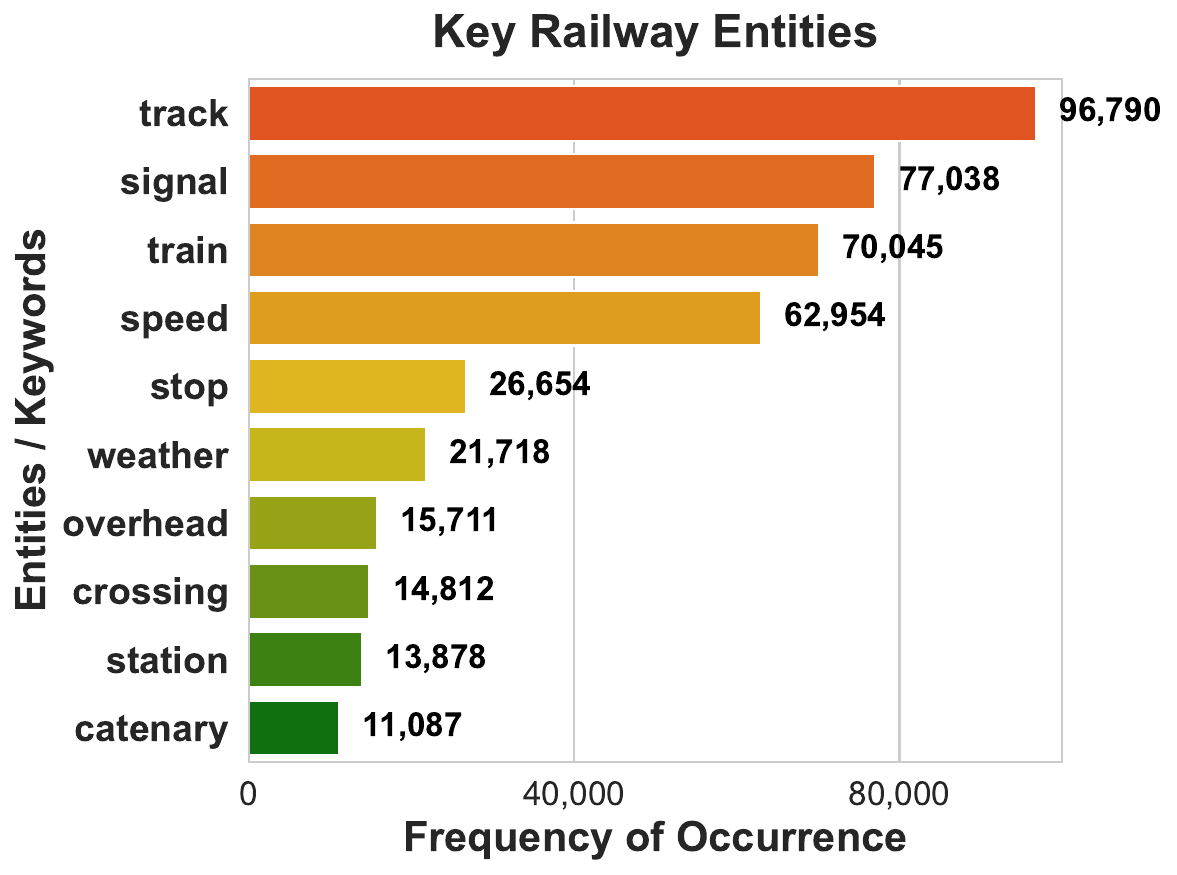}
        \label{fig:railvqa_entities}
    }
    \hfill
    \subfloat[Question Intents Distribution]{
        \includegraphics[width=0.29\linewidth]{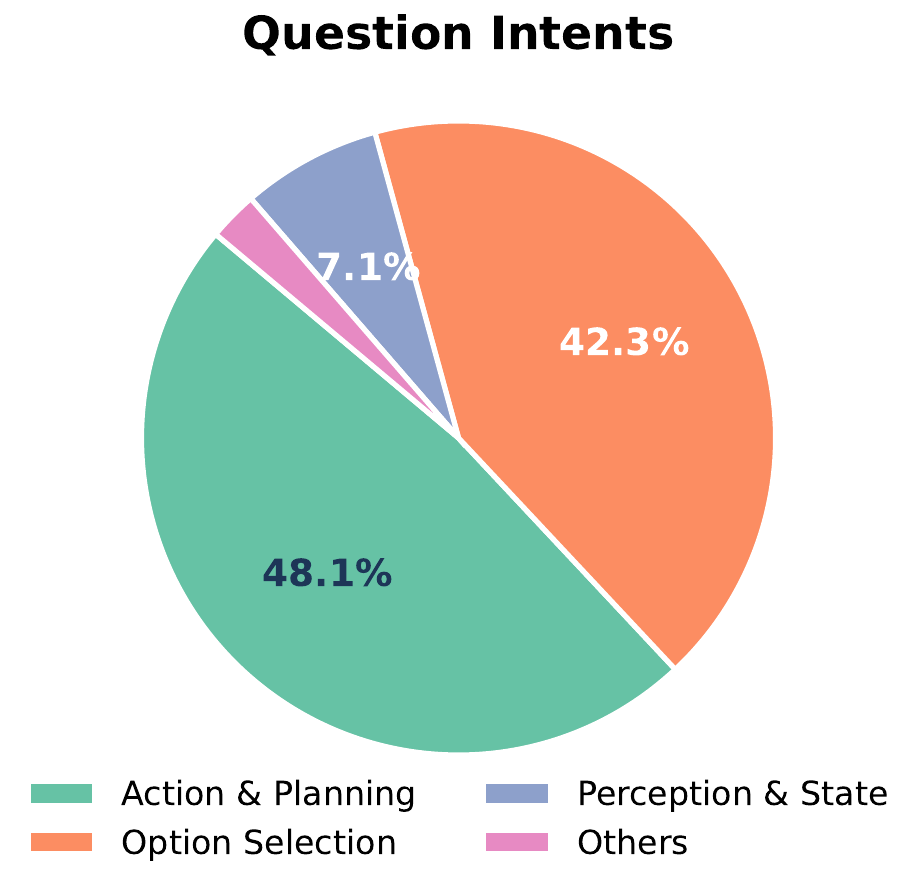}
        \label{fig:railvqa_intents}
    }
    
    \caption{\textbf{Comprehensive statistical overview of the RailVQA-bench dataset. (a) shows the distribution of generated CoT character lengths, reflecting the benchmark’s emphasis on logic-intensive reasoning. (b) reports the occurrence frequencies of key railway entities, demonstrating broad domain-specific coverage. (c) summarizes the distribution of question intents, indicating a primary focus on action planning and safety-related decisions rather than simple perception.}}
    \label{fig:railvqa_stats}
\end{figure*}

RailVQA-bench consists of two complementary subtasks—\textbf{Static Single-frame VQA} and \textbf{Dynamic Multi-frame VQA} (Fig.~\ref{fig:railvqa_main})—designed to evaluate cognitive reasoning in railway scenarios from a unified perspective. The static subtask mainly focus on rule compliance and right-of-way reasoning, while the dynamic subtask mainly assess kinematic risk awareness and safety-critical reasoning under temporal context. The statistical distributions of subtask-specific entities and question intents are reported in Figures~\ref{fig:railvqa_entities} and \ref{fig:railvqa_intents}, respectively.

\textbf{Single-frame VQA:} 
Given a single cab-view (first-person) image $I \in \mathbb{R}^{H \times W \times 3}$, the model must identify key visual elements and infer the appropriate operational command according to railway features. The main task is to accurately ground visual evidence in rules and regulatory constraints to produce a compliant operation suggestion.

\textbf{Multi-frame VQA:} 
The input video is a sequence of $T$ frames $S = \{I_1, I_2, ..., I_T\}$ with a question, and the model must track dynamic objects over time and discriminate benign background motion (e.g., passengers on a platform) from hazardous intrusions (e.g., pedestrians entering the tracks). It also requires physical commonsense reasoning that reflects the train’s high momentum and one-degree-of-freedom motion constraint, so as to be able to assess the risk based on the relative distance and speed, and support emergency decisions beyond single-frame recognition.

\textbf{Structured Chain-of-Thought (CoT) Output:} 
To improve interpretability and enable standardized evaluation, both tasks require responses in a structured chain-of-thought (CoT) format. Formally, given a visual input  $V$ (either a single-frame image $I$ or a frame sequence $S$) and  a  question $Q$, the model learns a function $\mathcal{F}$ that produces:

\begin{equation}
\mathcal{F}(V, Q) \rightarrow \{ \mathcal{C}_{perc}, \mathcal{C}_{reas}, \mathcal{C}_{plan}, \mathcal{A}_{final} \}
\end{equation}
where $\mathcal{C}_{perc}$ summarizes salient visual evidence (e.g., signal aspect), $\mathcal{C}_{reas}$ denotes logical/physical inference, $\mathcal{C}_{plan}$ represents decision-oriented inference, and $\mathcal{A}_{final}$ is the final concise answer. This structure enforces explicit reasoning stages and enables fine-grained evaluation of model.

\subsection{Dataset Collection}
To construct a comprehensive and diverse benchmark for railway visual reasoning, we build RailVQA-bench by integrating and strictly filtering two public datasets: RailSem19~\cite{zendel2019railsem19} and RailGoerl24~\cite{tagiew2025railgoerl24}. The collection protocol targets two core capabilities: (i) single-frame scene understanding and (ii) cross-frame temporal reasoning.

\textbf{Static Scenario Collection:} For single-frame asset recognition and hazard identification, we compile a large-scale static subset consisting of 8,500 images from RailSem19 and 1,500 high-quality frames selected from RailGoerl24. Together, they cover diverse environmental conditions (e.g., illumination and weather) and operational contexts (e.g., urban tramways and high-speed rail).

\textbf{Dynamic Scenario Collection:}  To evaluate temporal reasoning in train operations, we further process the video data in RailGoerl24. Using temporal segmentation and semantic re-labeling, we restructure the raw streams into 584 dynamic scenarios capturing safety-critical events such as intrusions and signal-state transitions, which form the basis of our cross-frame reasoning tasks.

\subsection{Automatic QA Generation Pipeline}
\label{sec:generation}
Following the automated benchmark construction paradigm of Space-LLaVA~\cite{foutter2024space}, we develop an automated QA generation pipeline using Qwen-VL-Max~\cite{bai2025qwen3} as the teacher model. It is selected for its fine-grained visual grounding and its robust long-context modeling for multi-frame temporal analysis. Through rigorous prompting, we generate explicit Chain-of-Thought (CoT) annotations and extract multiple queries per visual input. By creating one open-ended question and one multiple-choice question for each static image and each video clip, we systematically expand the raw 10,000 images and 584 video clips into 20,000 static QA pairs and 1,168 dynamic QA pairs.  See Appendix~\ref{app:generation_pipeline} for prompting details.

\subsection{Dataset Validation}
\label{sec:validation}
To ensure annotation quality, we adopted a validation protocol tailored to the two subtasks. Because multi-frame reasoning is difficult to generate and verify reliably, we manually reviewed the entire dynamic subset, correcting minor issues and removing samples with substantial errors. Given the larger scale and relative simplicity of the static subset, we conducted statistical quality control by auditing 1,000 instances. This audit yielded a pass rate exceeding 95\%, indicating high annotation fidelity and providing further evidence for the trustworthiness of our automated generation process for the static subset.

\subsection{Evaluation Methodology}
To assess high-level reasoning and decision-making on RailVQA-bench, we adapt the rubric-based protocol of DriveLMM-o1~\cite{ishaq2025drivelmm} - originally developed for autonomous driving - to the railway domain. The resulting framework defines 12 semantic dimensions (Table~\ref{tab:metrics_rubric}) and aims to evaluate the qualities of perception and reasoning in train-driving scenarios.

We further adopt an LLM-as-a-Judge~\cite{zheng2023judging} setting to evaluate the semantic quality of predicted CoTs. Specifically, we use Qwen3-Max~\cite{bai2023qwen} as the judge model to compare model outputs against manually revised references, leveraging its strong instruction following and logical reasoning. Prior work reports strong agreement between  large-model judges and human experts on NLP benchmarks~\cite{liu2023g}. Moreover, this paradigm has been increasingly adopted in recent top-tier studies on autonomous driving~\cite{sima2024drivelm, xu2024drivegpt4} and visual reasoning (VQA).

\begin{table*}[t]
\centering
\renewcommand{\arraystretch}{1.1}
\caption{Comparison of RailVQA-bench with related datasets. Existing datasets such as OSDaR23~\cite{tagiew2023osdar23} and SynRailObs~\cite{guo2025synrailobs} primarily advance perception (e.g., via multi-sensor fusion or synthetic data generation) but provide limited high-level cognitive annotations. RailVQA-bench addresses this limitation by explicitly coupling perception with reasoning and planning.}
\label{tab:dataset_comparison}
\resizebox{1.0\textwidth}{!}{%
\begin{tabular}{l|c|cc|ccc|l}
\toprule
\multirow{2}{*}{\textbf{Dataset}} & \multirow{2}{*}{\textbf{Year}} & \multicolumn{2}{c|}{\textbf{Modality}} & \multicolumn{3}{c|}{\textbf{Annotations}} & \multirow{2}{*}{\textbf{Task Focus}} \\
 &  & \textit{Static} & \textit{Dynamic} & \textit{Bbox/Mask} & \textit{Q\&A} & \textit{Logic (CoT)} &  \\ \midrule
RailSem19~\cite{zendel2019railsem19} & 2019 & \checkmark & - & \checkmark & - & - & Sem. Seg. \& Detection \\
OSDaR23~\cite{tagiew2023osdar23} & 2023 & \checkmark & \checkmark & \checkmark & - & - & Multi-sensor Perception \\
RailGoerl24~\cite{tagiew2025railgoerl24} & 2024 & - & \checkmark & \checkmark & - & - & Visual Tracking \\
SynRailObs~\cite{guo2025synrailobs} & 2025 & \checkmark & - & \checkmark & - & - & Synthetic Obstacle Detection \\ \midrule
DriveLM~\cite{sima2024drivelm}(Auto Driving) & 2023 & \checkmark & \checkmark & \checkmark & \checkmark & \checkmark & Vehicle Driving Reasoning \\ \midrule
\textbf{RailVQA-bench (Ours)} & \textbf{2026} & \textbf{\checkmark} & \textbf{\checkmark} & \textbf{\checkmark} & \textbf{\checkmark} & \textbf{\checkmark} & \textbf{Perception \& Reasoning} \\ \bottomrule
\end{tabular}%
}
\end{table*}

\section{RailVQA-CoM: Collaborative Large-Small Models Framework}

This section introduces RailVQA-CoM, a collaborative framework designed to better accommodate dynamic
multi-frame scenarios. In addition, to evaluate spatial understanding ability, we adopt a lightweight RailVQA-CoM variant that retains the small-model perception module and the LMM-based cognitive inference module for static scenarios. In both settings, the input is a query paired with either a video clip or a single image.

\subsection{Small Model Perception Module}
This module focuses on the rapid and precise acquisition of structured visual primitives from the raw video stream or a static image.

\textbf{Customized Railway Detector.}
To balance efficiency and domain accuracy, we adopt YOLO26m~\cite{redmon2016you} as the backbone detector. We fine-tune it on RailSem19 and extend the label space with additional manual annotations for common non-railway obstacles (e.g., pedestrians and vehicles), enhancing its generalization ability.

\textbf{State Estimation and Tracking.}
We employ ByteTrack~\cite{zhang2022bytetrack} for multi-object tracking. By associating detections across frames using IoU (Intersection over Union) and motion consistency, ByteTrack assigns each target a persistent identity $id$, ensuring cross-frame identity continuity.

\textbf{Output and Visual Enhancement.}
For each frame $t$, the module outputs a structured detection set:
\begin{equation}
B_t = \{ (x_1, y_1, x_2, y_2, c, id)_i \mid i \in \mathcal{O}_t \}
\end{equation}
where $(x_1,y_1,x_2,y_2)$ are bounding-box coordinates, $c$ is the semantic class, $id$ is the tracking identity, and $\mathcal{O}_t$ represents the set of all active tracked objects. We also add the sequence index (e.g., ``Seq: $t$/$T$'') on each frame to provide an explicit temporal anchor for the downstream LMM.

\subsection{Motion Analysis and Memory Log Module}

This module is designed specifically to handle dynamic multi-frame scenarios. It converts visual information into temporally coherent, physically meaningful semantics within a memory log.

\textbf{Multi-dimensional Motion Estimation.}
For each tracked target $i$, we compute an instantaneous 2D motion vector from its center $C_t^i=(x_c,y_c)_t$ over an interval $\Delta t$:
\begin{equation}
\mathbf{v}_t^{i} = \frac{C_t^i - C_{t-\Delta t}^i}{\Delta t} = (v_x, v_y)
\end{equation}

Furthermore, to capture depth-wise movement (e.g., objects approaching the train along the tracks), we introduce the \textbf{Area Change Rate} $\rho_t^i$. This rate is defined as
\begin{equation}
\rho_t^i = \frac{A_t^i - A_{t-\Delta t}^i}{A_{t-\Delta t}^i}
\end{equation}
where $A_t^i$ is the area calculated by bounding box in frame $t$.

\textbf{Kinematic State Inference.}
To filter out perceptual noise while retaining sensitivity to critical hazards, we employ a dual-threshold gating mechanism.
Besides area change rate, we use an adaptive threshold conditioned on the bounding-box width $w^i$:
\begin{equation}
\tau_{dyn} = \max(\tau_{min}, \lambda \cdot w^i)
\end{equation}
where $\tau_{min}$ is a baseline noise floor and $\lambda$ controls scale adaptation. The kinematic state is defined as
\begin{equation}
State_{t}^{i}=
\begin{cases}
Moving,& \text{if } ||\mathbf{v}_{t}^{i}|| \geq \tau_{dyn} \text{ or } |\rho_t^i| > \gamma \\ Static,&otherwise
\end{cases}
\end{equation}

A target is classified as Moving if either its spatial displacement exceeds the dynamic threshold $\tau_{dyn}$ or its absolute area change rate $|\rho_t^i|$ surpasses a sensitivity constant $\gamma$.

\textbf{Identity Track Maintenance.}
To address track fragmentation caused by transient occlusions or detector false negatives, we introduce a robust \textbf{Spatio-temporal ID} mechanism. This approach actively re-associates newly detected, unmatched objects (denoted as $d_j$) with recently lost identities (denoted as $i^*$) by evaluating a joint spatio-temporal constraint $\mathcal{C}$. Here is the explanation of constraint $\mathcal{C}$:
\begin{equation}
\mathcal{C}(d_j, i^*) \iff 
\begin{cases}
Class(d_j) = Class(i^*), \text{ and} \\
t - t_{last}^{i^*} \le \delta_{tol}, \text{ and} \\
||C_t^j - C_{last}^{i^*}||_2 < \theta
\end{cases}
\end{equation}

When a target disappears, it is maintained in a short-term buffer for a tolerance period $\delta_{tol}$. If a new detection of the same class appears within this window, identity recovery is triggered based on a scale-aware spatial search radius $\theta=\max(2 \cdot w^j, \lambda_{min})$. This dynamic association ensures temporal continuity across complex scenes.

\subsection{Large Model Cognitive Inference Module.}
We introduce three mechanisms for efficient and interpretable LMM training-free inference:

\textbf{Dynamic Budget Allocation and Adaptive Sampling.}
To mitigate computational overhead and context dilution, we applied an event-driven dynamic budget allocation mechanism. For a sequence of length $T$, we first compute a heuristic event score $s_t$ for each frame based on the intensity of dynamic events (e.g., object appearances, significant spatial displacements). The overall scene complexity is aggregated as $\mathcal{S} = \sum_{t=1}^{T} s_t$.

The system dynamically assigns an effective frame budget $K \le K_{max}$ based on the scene complexity:
\begin{equation}
K = \begin{cases} 
K_{max}, & \text{if } \mathcal{S} \ge \tau_{high} \\ 
\max(3, \lfloor \alpha \cdot K_{max} \rfloor), & \text{if } \tau_{low} \le \mathcal{S} < \tau_{high} \\ 
2, & \text{if } \mathcal{S} < \tau_{low} 
\end{cases}
\end{equation}
where $\tau_{low}$ and $\tau_{high}$ are complexity thresholds, dynamically set based on the complexity score distribution calculated on a calibration set using our perception and motion analysis mechanisms. This mechanism facilitates easy threshold adjustment across different datasets or scenarios. And $\alpha$ is a scaling factor.

Finally, to extract the most informative moments, the sequence is divided into $K-1$ temporal segments. The frame $t_k^*$ with the highest local score $s_t$ within each segment is selected as a keyframe. The final set $\mathcal{T}^* = \{t_1^*, \dots, t_{K-1}^*, T\}$ includes the last frame to anchor the current visual state.

\textbf{Multi-modal Fusion and Defensive Prompting.}
We construct the multi-modal prompt by concatenating the keyframes $\mathcal{T}^*$ with the structured textual logs and query. Yet, because introduced lightweight perception detectors inherently struggle with out-of-distribution (OOD) anomalies, we implement a Defensive Prompting mechanism. When the log consecutively outputs a ``No objects'' state, the system dynamically injects a fallback instruction, forcing the LMM to disregard the empty log and rely exclusively on its own visual comprehension to detect atypical hazards.

\textbf{Structured Chain-of-Thought (CoT) Generation.}
Finally, we embed a structured Chain-of-Thought (CoT) template into the system prompt. This instruction guides the LMM to articulate its perceiving-reasoning-planning steps, enhancing the interpretability and transparency of system response.

\section{Experiments}

\subsection{Experiment Setup}

\textbf{Baselines and Evaluation Setting.} To demonstrate the model-agnostic properties of our proposed framework, we selected four mainstream open-source LMMs with varying parameter scales: Qwen3-VL-Instruct-8B~\cite{bai2025qwen3}, InternVL3.5-8B~\cite{chen2024internvl}, LLaVA-1.5-7B~\cite{liu2023visual}, LLaMA3-vision-11B~\cite{chi2024llama}.

All experiments are conducted on RailVQA-bench  instead of  conventional railway datasets (e.g., RailSem19~\cite{zendel2019railsem19} and OSDaR23~\cite{tagiew2023osdar23}), as the latter are designed mainly for low-level perception tasks, thus do not provide the QA supervision or reasoning annotations.

To unify the evaluation of the overall score, the ratings for \textit{Hallucination}, \textit{Missing Step}, \textit{Missing details} are scaled such that higher scores represent better performance. Please refer to Table~\ref{tab:metrics_rubric} in Appendix for specific scoring details.

We consider two baseline configurations for comparison:

1) \textbf{LMM Only}: The LMM model answers user queries directly from raw static images or video frames, relying solely on its pre-trained zero-shot capability without external perceptual priors.

2) \textbf{Unadapted Detector}: This setting preserves the full temporal pipeline but replaces the perception module with a standard, unfine-tuned YOLO26m detector, isolating the contribution of domain adaptation in the perception module from that of the reasoning middleware.

3) \textbf{Uniform Samples}: This configuration utilizes the complete RailVQA-CoM framework but disables the dynamic budget allocation. It equidistantly extracts a fixed number (equal to the maximum frame of the dynamic sampling) of frames across the temporal sequence.

4) \textbf{RailVQA-CoM (Ours)}: Our proposed framework, which integrates domain-adapted YOLO26m-Rail perception to support the LMM’s cognitive reasoning.

\textbf{Implementation Details.} 
Experiments were conducted with four NVIDIA RTX A6000 GPUs. We fine-tuned only the YOLO26m~\cite{redmon2016you} detector on our generalization-enhanced railway dataset, while keeping the LMM frozen to preserve generality. For evaluation, we used Qwen3-Max to score the generated chain-of-thought (CoT) outputs across the 12 semantic dimensions defined in RailVQA-bench.

\subsection{Dynamic Scene Inference Experiments.}
\begin{table*}[htbp]
\centering
\renewcommand{\arraystretch}{1.2}
\caption{Quantitative comparison on the Dynamic Scene VQA task. ($\uparrow$) indicates higher is better, where the criteria of scoring are listed in appendix. Note that System TPS measures the effective token generation rate over the entire system.}
\label{tab:dynamic_inference}
\resizebox{\textwidth}{!}{%
\begin{tabular}{llccccccccc}
\toprule
\textbf{LMM} & \textbf{Method} & \textbf{CQ Acc. (\%)} & \textbf{Overall Score} & \textbf{Physics \& Mom. ($\uparrow$)} & \textbf{Risk Assess. ($\uparrow$)} & \textbf{Rule Adhere. ($\uparrow$)} & \textbf{Details ($\uparrow$)} & \textbf{Hallucination ($\uparrow$)} & \textbf{S-TPS ($\uparrow$)} \\ 
\midrule
\multirow{3}{*}{\textbf{InternVL3.5}}
 & LMM Only   & 67.16 & 60.60 & 69.4 & 54.4 & 71.6 & 53.0 & 53.8 & 5.80 \\
 & Unadapted Detector & 60.45 & 52.50 & 67.6 & 41.1 & 71.5 & 43.9 & 51.6 & 13.32 \\
 & Uniform Samples & 74.80 & 66.73 & 75.2 & \textbf{58.8} & \textbf{80.1} & \textbf{59.6} & 61.0 & 11.54 \\
 & \textbf{RailVQA-CoM} & \textbf{75.43} & \textbf{67.92} & \textbf{75.3} & 55.7 & 79.9 & 59.2 & \textbf{71.4} & \textbf{17.06} \\
\midrule
\multirow{3}{*}{\textbf{Qwen3-VL}}
 & LMM Only   & 84.36 & 71.01 & 78.9 & 63.8 & 79.6 & 54.8 & 71.5 & 6.04 \\
 & Unadapted Detector & 82.17 & 71.80 & 81.3 & 64.7 & 85.4 & 58.9 & 71.0 & 9.25 \\
 & Uniform Samples & 82.53 & 74.00 & 80.6 & \textbf{68.4} & 84.8 & 66.5 & 74.4 & 8.16 \\
 & \textbf{RailVQA-CoM} & \textbf{84.93} & \textbf{76.64} & \textbf{81.8} & 68.1 & \textbf{86.4} & \textbf{67.4} & \textbf{81.2} & \textbf{10.34} \\
\bottomrule
\end{tabular}%
}
\end{table*}

\begin{table*}[htbp]
\centering
\renewcommand{\arraystretch}{1.2}
\caption{Quantitative comparison on the Static Scene VQA task. We report a comprehensive suite of fine-grained metrics to quantify cognitive improvements. ($\uparrow$) indicates higher is better.}
\label{tab:static_inference}
\resizebox{\textwidth}{!}{%
\begin{tabular}{llcccccccc}
\toprule
\textbf{LMM} & \textbf{Method} & \textbf{Overall Score} & \textbf{CQ Acc. (\%)} & \textbf{Faithfulness ($\uparrow$)} & \textbf{Risk Assess. ($\uparrow$)} & \textbf{Rule Adherence ($\uparrow$)} & \textbf{Object Underst. ($\uparrow$)} & \textbf{Hallucination ($\uparrow$)} & \textbf{Details ($\uparrow$)} \\ 
\midrule
\multirow{3}{*}{\textbf{LLaVA-1.5}}   
 & LMM Only    & 69.82 & 72.20 & 68.1 & 69.1 & 70.9 & 67.7 & 76.1 & 60.8 \\
 & Unadapted Detector  & 71.10 & 75.81 & 69.0 & 70.0 & 71.7 & 69.1 & \textbf{79.0} & 61.8 \\
 & \textbf{RailVQA-CoM} & \textbf{71.48} & \textbf{74.90} & \textbf{70.0} & \textbf{70.5} & \textbf{72.5} & \textbf{69.7} & 78.8 & \textbf{63.0} \\ 
\midrule
\multirow{3}{*}{\textbf{LLaMA3-Vision}} 
 & LMM Only    & 65.72 & \textbf{81.68} & \textbf{67.2} & 65.4 & 67.2 & 62.1 & 70.5 & 57.7 \\
 & Unadapted Detector  & 66.57 & 80.56 & 65.5 & 66.3 & 68.5 & \textbf{63.5} & 73.4 & 58.2 \\
 & \textbf{RailVQA-CoM} & \textbf{67.00} & 80.74 & 63.8 & \textbf{66.7} & \textbf{70.5} & 63.0 & \textbf{74.2} & \textbf{58.6} \\ 
\midrule
\multirow{3}{*}{\textbf{InternVL3.5}} 
 & LMM Only    & 71.04 & 78.10 & 74.5 & 73.6 & 75.1 & 73.3 & 71.7 & 61.0 \\
 & Unadapted Detector  & 73.20 & 81.55 & 74.4 & 71.7 & 76.6 & 68.9 & 79.5 & 62.4 \\
 & \textbf{RailVQA-CoM} & \textbf{74.08} & \textbf{82.40} & \textbf{76.5} & \textbf{75.8} & \textbf{77.0} & \textbf{74.4} & \textbf{82.1} & \textbf{65.2} \\ 
\midrule
\multirow{3}{*}{\textbf{Qwen3-VL}}    
 & LMM Only    & 74.81 & 85.20 & \textbf{76.3} & 74.9 & 78.2 & 71.2 & 81.1 & 64.9 \\
 & Unadapted Detector  & 72.20 & 86.10 & 70.7 & 73.0 & 75.5 & 68.1 & 81.7 & 60.1 \\
 & \textbf{RailVQA-CoM} & \textbf{75.74} & \textbf{86.40} & 74.3 & \textbf{75.3} & \textbf{79.3} & \textbf{72.8} & \textbf{87.9} & \textbf{67.8} \\ 
\bottomrule
\end{tabular}%
}
\end{table*}

\textbf{Experimental Setting} 
To systematically evaluate our proposed framework in multi-frame (video clip) reasoning and efficiency, we introduce system-level metric: \textit{System TPS (S-TPS)}~\cite{zhou2024survey}, which is defined as the total number of generated tokens divided by the end-to-end latency of the entire large–small model framework.

\begin{figure}[htbp]
    \centering
    \includegraphics[width=0.835\columnwidth]{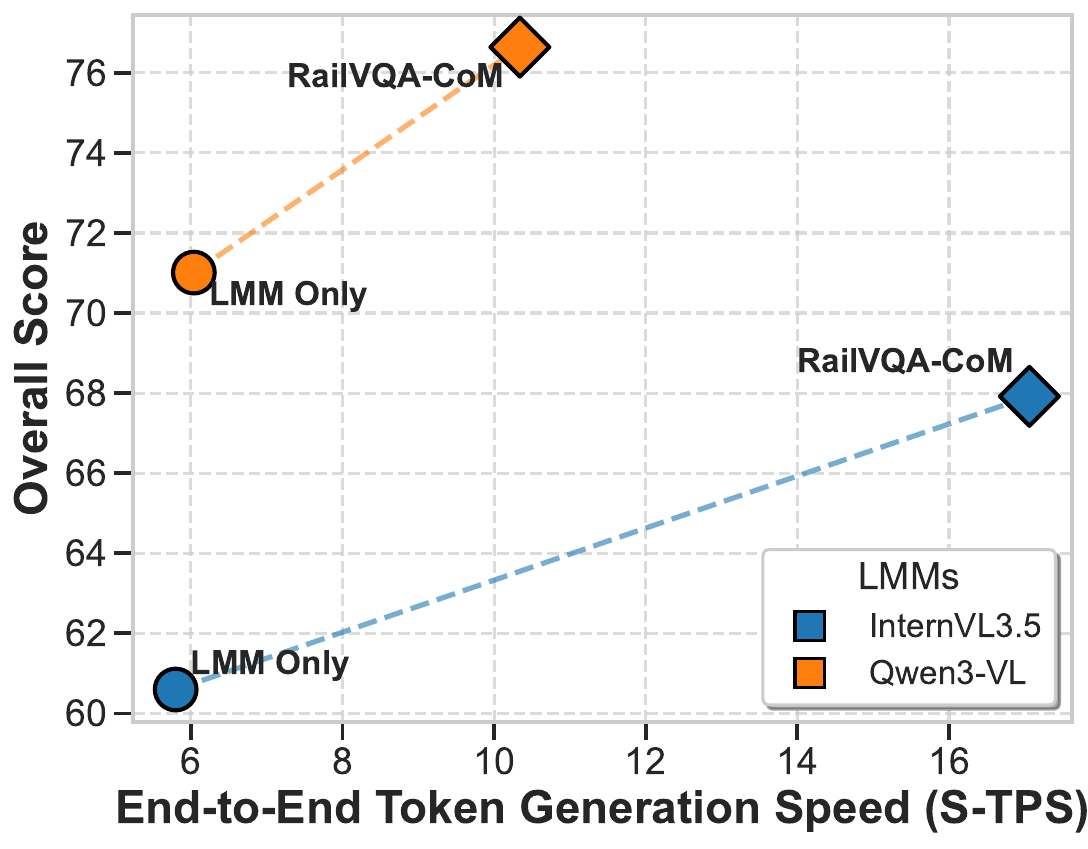}
    \caption{Performance-efficiency comparison in dynamic scenarios. RailVQA-CoM achieves simultaneous substantial gains, pushing the performance towards the top-right by increasing end-to-end throughput and enhancing cognitive scores.}
    \label{fig:performance}
\end{figure}

\textbf{Results and Analysis.}
As visually demonstrated in Fig.~\ref{fig:performance}, RailVQA-CoM successfully improves the performance-efficiency ability of the LMM Only baseline, propelling both models from the suboptimal lower-left toward the optimal top-right quadrant. Table~\ref{tab:dynamic_inference} quantitatively confirms this dual breakthrough. On the efficiency front, adaptive keyframe sampling effectively increases the LMM reasoning throughput, for example InternVL3.5's QA speed surges from 5.80 to 17.06 S-TPS, which has even increased by three times. On the cognitive front, the results demonstrate profound improvements in dynamic hazard anticipation, driving Physics \& Momentum score from 69.4 to 75.3 in InternVL3.5, and Hallucination Score from 53.8 to 71.4.
Overall, compared to traditional approaches, our framework demonstrates a capability improvement in processing continuous dynamic visual streams.

\subsection{Static Scene Inference Experiments}

\textbf{Experimental Setting.}
To evaluate spatial understanding and rule compliance in static frame, we apply a lightweight Visual Prompting mechanism for the RailVQA-CoM and Unadapted Detector configurations, bypassing the temporal modules. Specifically, bounding boxes and semantic labels from the perception module are rendered onto the RGB image. This augmented image, alongside a customized Chain-of-Thought prompt, is then fed into the LMM.

\textbf{Results and Analysis.}
As shown in Table~\ref{tab:static_inference}, RailVQA-CoM consistently achieves the highest Overall Score and CQ Accuracy across all baselines.

Comparing the LMM Only with the Unadapted Detector demonstrates the fundamental benefit of RailVQA-CoM framework. Explicitly overlaying bounding boxes successfully anchors the LMM's attention, significantly reducing hallucinations (e.g., the Hallucination score of InternVL3.5 jumps from 71.7 to 79.5). However, the un-finetuned Unadapted Detector struggles with railway-specific assets. Its generic bounding boxes can mislead the LMM, causing a noticeable drop in specialized metrics like Object Understanding (e.g., from 71.2 down to 68.1 for Qwen3-VL).

RailVQA-CoM resolves this bottleneck by deploying the domain-adapted YOLO26m-Rail. This confirms that domain-specific visual cues are essential for reliable LMM reasoning in railway scenarios.

\subsection{Cross-domain Generalization}

\begin{table*}[htbp]
\centering
\renewcommand{\arraystretch}{1.2}
\caption{Cross-domain generalization on the DriveLMM-o1 benchmark for autonomous driving.}
\label{tab:cross_domain}
\resizebox{\textwidth}{!}{%
\begin{tabular}{llcccccccc}
\toprule
\textbf{Method} & \textbf{LMM} & \textbf{Overall Score} & \textbf{CQ Acc. (\%)} & \textbf{Faithfulness ($\uparrow$)} & \textbf{Risk Assess. ($\uparrow$)} & \textbf{Rule Adhere. ($\uparrow$)} & \textbf{Scene Aware. ($\uparrow$)} & \textbf{Hallucination ($\uparrow$)} & \textbf{Details ($\uparrow$)} \\ 
\midrule
\multirow{2}{*}{DriveLMM-o1} 
 & InternVL-2.5 & 57.48 & 54.11 & \textbf{49.84} & 52.55 & 67.22 & 54.67 & 51.20 & 51.69 \\
 & Qwen2.5-VL   & 56.63 & 51.14 & 47.43 & \textbf{53.08} & \textbf{68.52} & 49.91 & 50.96 & 51.70 \\
\midrule
\textbf{RailVQA-CoM} 
 & DriveLMM-o1(Qwen2.5-VL)   & \textbf{58.89} & \textbf{54.57} & 48.66 & 52.88 & 68.07 & \textbf{56.48} & \textbf{52.81} & \textbf{53.77} \\ 
\bottomrule
\end{tabular}%
}
\end{table*}

\begin{table*}[htbp]
\centering
\renewcommand{\arraystretch}{1.2}
\caption{Ablation study of core components.}
\label{tab:ablation}
\resizebox{\textwidth}{!}{%
\begin{tabular}{lcccccccc}
\toprule
\textbf{Variant} & \textbf{Overall Score} & \textbf{CQ Acc. (\%)} & \textbf{Risk Assess. ($\uparrow$)} & \textbf{Physics \& Mom. ($\uparrow$)} & \textbf{Object Underst. ($\uparrow$)} & \textbf{Semantic Cov. ($\uparrow$)} & \textbf{Hallucination ($\uparrow$)} & \textbf{Details ($\uparrow$)} \\ 
\midrule
w/o adaptive sampling   & 74.00 & 82.53 & 68.4 & 80.6 & 72.1 & 63.8 & 74.3 & 66.5 \\
w/o event stream log  & 64.67 & 78.80 & 56.5 & 76.9 & 58.3 & 56.6 & 60.3 & 57.2 \\
\textbf{Full RailVQA-CoM}   & \textbf{76.64} & \textbf{84.93} & \textbf{68.1} & \textbf{81.8} & \textbf{74.3} & \textbf{71.0} & \textbf{81.2} & \textbf{67.4} \\ 
\bottomrule
\end{tabular}%
}
\end{table*}

To verify that our collaborative paradigm extends beyond the railway domain, we evaluated its generalization on the autonomous vehicle driving benchmark, DriveLMM-o1~\cite{ishaq2025drivelmm}. The original DriveLMM-o1 method utilizes domain-specific models, where the InternVL-2.5 variant reported higher scores than the Qwen2.5-VL variant on their specific benchmark. To demonstrate the effectiveness of our framework, we deliberately wrapped the lower-scoring Qwen2.5-VL model within RailVQA-CoM to provide it with explicit perceptual priors.

As shown in Table~\ref{tab:cross_domain}, the integration yields compelling improvements. Empowered by RailVQA-CoM, the Qwen2.5-VL model not only achieves a significant performance leap over its original baseline, but also completely surpasses the stronger InternVL-2.5 expert model. Notably, its Scene Awareness score surges from 49.91 to 56.48, lifting the Overall Score to 58.89. This confirms that our proposed paradigm serves as a versatile and LMM training-free framework that can integrate with and enhance the performance of LMMs across diverse autonomous driving scenarios.

\subsection{Ablation Study}

To validate the necessity of \textit{Event Stream Log} and \textit{Adaptive Temporal Sampling} in our dynamic design, we conduct an ablation study on Qwen3-VL-8B. This classical model was chosen to represent our framework's capabilities; identical performance degradation trends were observed across the other baselines when these modules were removed.

\begin{figure}[htbp]
    \centering
    \includegraphics[width=\columnwidth]{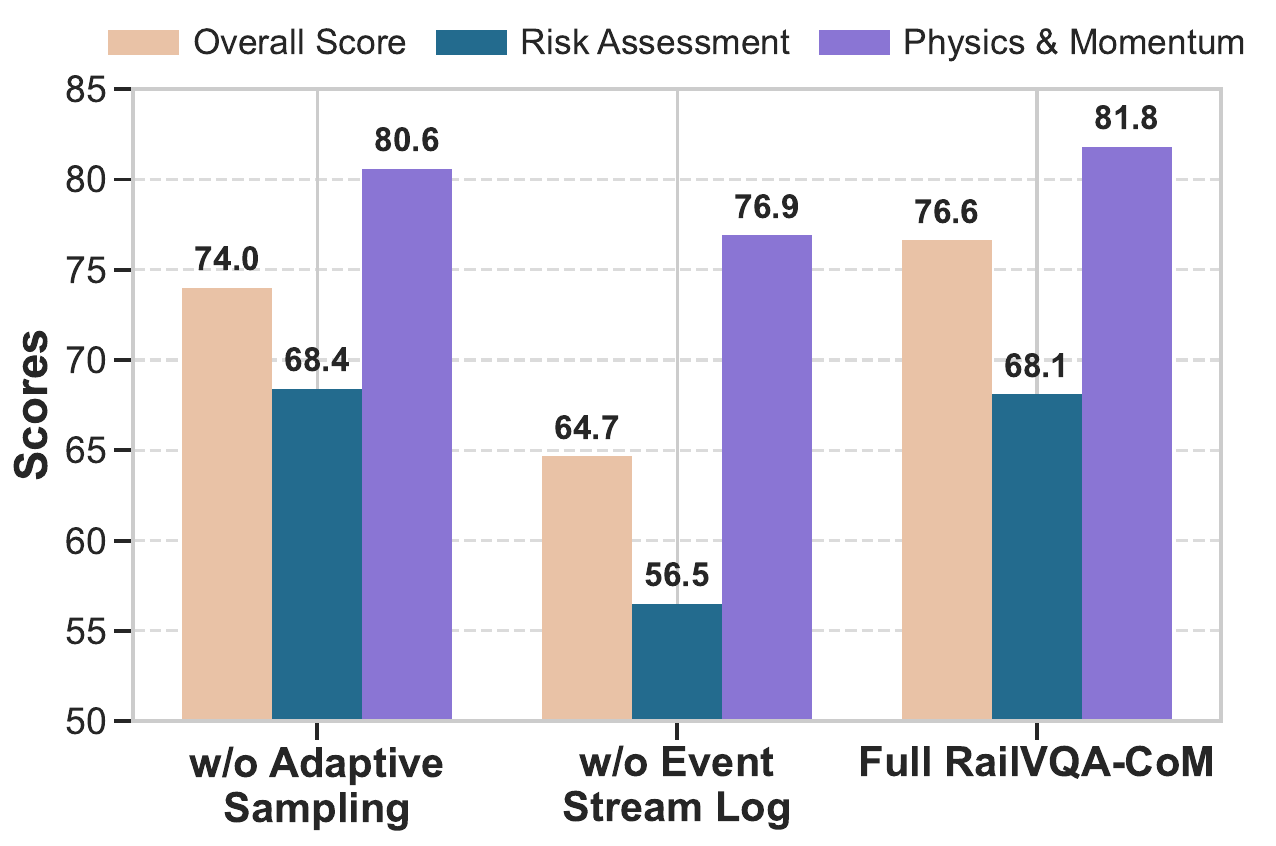}
    \caption{Ablation study of the core middleware components of RailVQA-CoM. To intuitively demonstrate the impact of each module on model performance, we extract three representative metrics (Overall Score, Risk Assessment, and Physics \& Momentum) from the ablation table to construct this bar chart.}
    \label{fig:ablation}
\end{figure}

Fig.~\ref{fig:ablation} and Table~\ref{tab:ablation} show that both components are critical. Removing the \textit{Event Stream Log} reduces the Overall Score to 64.67, largely due to sharp drops in \textit{Operational Risk Assessment} (56.5). Without explicit velocity cues, the model fails to maintain temporal continuity and cannot reliably anticipate collision risk. Replacing \textit{Adaptive Sampling} with uniform sampling also degrades the \textit{Missing Details} score, indicating that motion-saliency-driven selection is important for capturing key hazard moments.

\subsection{Qualitative Results}
\label{sec:qualitative_results}

\begin{figure*}[!t]
    \centering
    \begin{minipage}{0.49\linewidth}
        \centering
        \includegraphics[width=\linewidth]{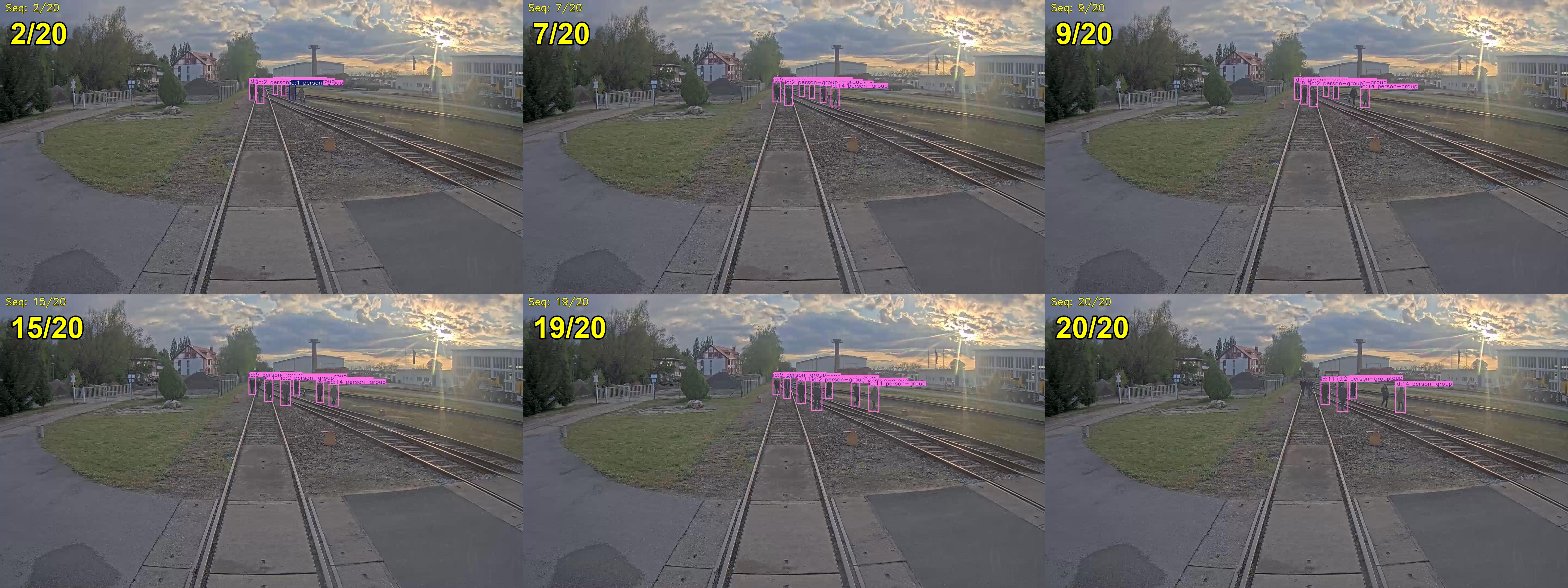}      
        \vspace{0.5em}
        \small (a) Case 1: Dynamic Intrusion with Multi-Object Tracking.
    \end{minipage}
    \hfill
    \begin{minipage}{0.49\linewidth}
        \centering
        \includegraphics[width=\linewidth]{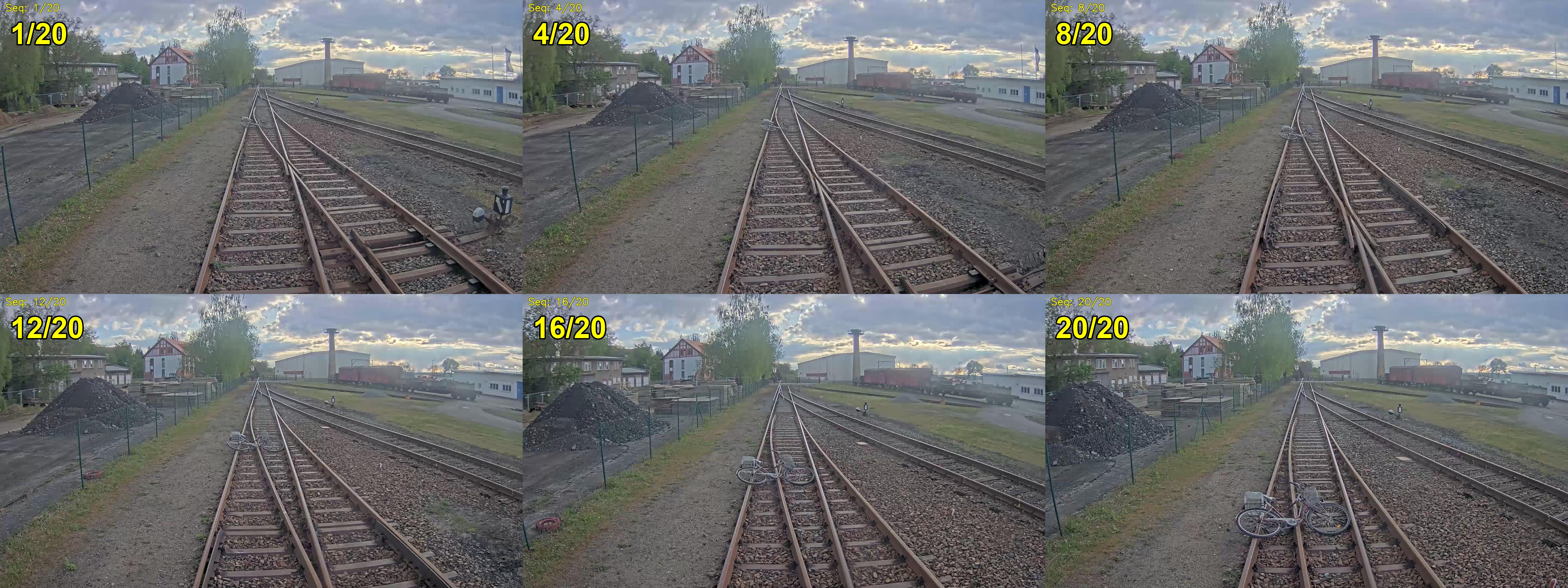}
        \vspace{0.5em}
        \small (b) Case 2: Zero-Shot Anomaly Detection (Defensive Fallback).
    \end{minipage}
    \vspace{1em} 
    \caption{Qualitative results of the RailVQA-CoM framework. (a) Case 1: The perception module tracks pedestrians and handles transient occlusions via track maintenance, preserving spatial-temporal context for the LMM. (b) Case 2: When the detector misses an out-of-distribution obstacle (bicycle), the system triggers a defensive fallback, prompting the LMM to identify the hazard via zero-shot visual reasoning.}
    \label{fig:qualitative_cases}
\end{figure*}

We provide two case studies (Fig.~\ref{fig:qualitative_cases}) to illustrate how RailVQA-CoM produces transparent, step-by-step reasoning and mitigates the impact of worthless inputs.

\vspace{0.5em}
\noindent \textbf{1) Case 1: Dynamic Intrusion and Track Maintenance}
\vspace{1ex}

As illustrated in Fig. \ref{fig:qualitative_cases}(a), the small detector captures the targets, and our motion analysis module robustly handles transient tracking losses via track maintenance. As shown in the generated logs below, temporarily occluded targets are successfully preserved in the event stream with \texttt{[Occluded/Lost]} descriptors, maintaining a complete spatial-temporal stream.

\vspace{0.5em}
\noindent \textbf{Generated Perception Log (Middleware Output Snippet):}
\begin{quote}
\ttfamily \scriptsize
... \\
Frame 15: person (ID:4) at [40,230,112,433] ... Motion: moving left | ... | person (ID:1)  [Occluded/Lost] predicted at approx [1916, 1024] \\
Frame 16 to Frame 20: person (ID:4) at [31,229,109,433] ... Motion: moving left | ... | person (ID:1) [Occluded/Lost] predicted at approx [1428, 565]
\end{quote}

\vspace{0.5em}
\noindent \textbf{RailVQA-CoM Output (LMM Structured CoT):}
\begin{quote}
\itshape \small
\textbf{Question:} What dynamic trend is observed in the pedestrian movement across the railway tracks?\\
\textbf{Perceiving:} Multiple individuals are moving along the railway tracks. The perception log confirms their motion is consistently rightward and slightly downward.\\
\textbf{Reasoning:} The motion vectors in the log indicate a clear, continuous movement away from the camera's viewpoint. Track intrusion poses an extreme collision risk.\\
\textbf{Planning:} Immediate action required for train: activate emergency brakes/signals and alert railway personnel.\\
\textbf{Final Answer:} The pedestrians are moving in a continuous, coordinated flow from left to right across the railway tracks.
\end{quote}

\vspace{0.5em}
\noindent \textbf{2) Case 2: Zero-Shot Anomaly Detection}
\vspace{1ex}

This is an extreme edge case, shown in Fig. \ref{fig:qualitative_cases}(b), rare in actual ATO operations, which serves as a rigorous stress test for handling out-of-distribution (OOD) anomalies by the RailVQA-CoM framework.

\vspace{0.5em}
\noindent \textbf{Triggered Middleware Status (Perception Log + Warning):}
\begin{quote}
\ttfamily \scriptsize
Frame 1: No objects. \\
Frame 2: No objects.\\
...\\
Frame 20: No objects.\\
\\
--- AUTOMATED DETECTOR STATUS: UNRELIABLE ---\\
System Report: The automated object detector found NO standard targets (People/Cars).\\
CRITICAL WARNING: The detector often FAILS to see non-standard obstacles (e.g., Bicycles, Rocks, Debris).\\
INSTRUCTION: IGNORE the detector's claim of 'No objects'. You MUST rely SOLELY on the Visual Images to find anomalies.
\end{quote}

\vspace{0.5em}
Under this fallback mechanism, the LMM disregards the unreliable log and instead relies on its inherent zero-shot visual grounding to identify the obstacles, thereby inferring the imminent collision hazard and the consequent derailment risk.

\vspace{0.5em}
\noindent \textbf{RailVQA-CoM Output (LMM Structured CoT):}
\begin{quote}
\itshape \small
\textbf{Question:} What dynamic event is occurring on the railway tracks in the sequence?\\
\textbf{Perceiving:} The visual sequence shows a bicycle lying across the left rail of the tracks, positioned in the path of travel.\\
\textbf{Reasoning:} As the train progresses along the tracks, it will inevitably encounter the stationary bicycle lying across the rails. This presents a severe and immediate hazard, as the bicycle will be struck, posing a significant risk of the train.\\
\textbf{Planning:} The system must flag this as a critical safety violation requiring immediate intervention to remove the obstacles from the tracks.\\
\textbf{Final Answer:} A stationary bicycle is illegally placed on the active railway tracks, creating a severe safety hazard.
\end{quote}

\section{Conclusion}

We present RailVQA-bench, the first VQA benchmark to comprehensively evaluate both static and dynamic, safety-critical kinematic reasoning for cab-view visual cognition in ATO, and RailVQA-CoM, an efficient large–small collaborative framework that supports a model-agnostic architecture for LMM deployment. By decoupling high-frequency perception from complex reasoning, our approach simultaneously improves computational efficiency and cognitive accuracy for LMMs. In critical dynamic scenarios, it increases end-to-end throughput per second about 3 times, while effectively overcoming context dilution and hallucinated reasoning. Overall, this work preliminarily advances reliable and efficient cognitive intelligence for railway automation.

\section{Limitations}

Although our framework substantially improves visual reasoning, it currently relies solely on monocular vision and does not incorporate multi-sensor fusion (e.g., LiDAR or millimeter-wave radar). Such fusion remains essential for accurate 3D depth estimation and robust all-weather perception. Future work will focus on integrating complementary sensing modalities to address these limitations.

Furthermore, it is worth noting that current Large Multi-modal Models still exhibit certain latency limitations. Although our collaborative framework brings noticeable efficiency improvements, directly deploying such large-scale models in Automatic Train Operation may still face practical latency constraints. We believe that future advances in LMMs will help alleviate this problem, and we also plan to investigate model acceleration techniques to further mitigate it.


\appendix[Prompt Engineering Details]
\label{sec:prompt_details}

To ensure the reproducibility of our methodology and provide transparency into the behavior of the Large Multi-modal Model (LMM), we report the core prompts used in the three main phases of our study: dataset generation, collaborative inference, and automated evaluation.

\section*{\textbf{(I) Prompt for Automatic QA Generation}}
\label{app:generation_pipeline}

To construct high-quality and domain-specific logical Question-Answering pairs for RailVQA-bench. We employed a strict system prompt to establish a professional persona and a structured user prompt to enforce the output format, constraining the output to a predefined JSON schema.

\vspace{0.5em}
\noindent \textbf{System Prompt (Persona Definition):}
\begin{quote}
\itshape
You are a Senior Railway Operation Expert and Instructor. Your task is to analyze images from the cab view and generate professional question-answering data for an automatic train operation system.
\end{quote}

\vspace{0.5em}
\noindent \textbf{User Prompt (Task and Structural Constraint):}
\begin{quote}
\itshape
Analyze the provided train cab-view image. Generate one Question-Answer (QA) pair and one Choice Question (CQ).

Strictly output valid JSON with no Markdown formatting (do not use json). Use the following structure:
\normalfont
\begin{verbatim}
{
  "cot_perception": "Visual analysis: 
    Identify signals (aspect/color)...",
  "cot_reasoning": "Logical analysis: 
    Interpret the visual data based...",
  "cot_planning": "Action plan: 
    Determine the immediate driving...",
  "qa_question": "A critical, 
    scenario-specific question...",
  "qa_answer": "A detailed answer 
    based on the analysis.",
  "mc_question": "A multiple-choice 
    question focusing on specific...",
  "mc_options": {
    "A": "Option text", 
    "B": "Option text", ...
  },
  "mc_correct": "The correct option 
    letter (e.g., 'A')"
}
\end{verbatim}
\end{quote}

\begin{table*}[htbp] 
\centering
\small 
\caption{Detailed scoring rubric for the 12 evaluation metrics (score range: 1–10). In the examples below, \textbf{Excellent} corresponds to scores of 9–10, whereas \textbf{Poor} corresponds to scores of 1–2.}
\label{tab:metrics_rubric}
\renewcommand{\arraystretch}{1.3} 
\begin{tabular}{p{0.21\textwidth} p{0.35\textwidth} p{0.40\textwidth}}
\toprule
\textbf{Metric} & \textbf{Evaluation Focus} & \textbf{Scoring Criteria Examples (Excellent vs. Poor)} \\
\midrule
\textbf{1. Faithfulness-Step} & Alignment with Ground Truth and Standard Operating Procedures (SOPs). & \textbf{Excellent:} All steps correctly match reference SOPs. \newline \textbf{Poor:} Majority of steps contradict ground truth.\\
\textbf{2. Informativeness-Step} & Completeness of reasoning regarding train status and environment. & \textbf{Excellent:} Captures all critical info (Signals, Switches). \newline \textbf{Poor:} Poor extraction of relevant reasoning. \\
\textbf{3. Operational Risk Assess.} & Prioritization of high-risk hazards (distinguishing safe surroundings vs. intrusions). & \textbf{Excellent:} Prioritizes Emergency Braking for intrusions. \newline \textbf{Poor:} Misses obvious obstructions or critical signals. \\
\textbf{4. Signal \& Rule Adhere.} & Compliance with Railway General Operating Rules and Signal Systems. & \textbf{Excellent:} Fully compliant with signal aspects. \newline \textbf{Poor:} Promotes highly unsafe behavior (e.g., SPAD). \\
\textbf{5. Object Understanding} & Interpretation of railway assets and spatial location of dynamic objects. & \textbf{Excellent:} Correctly distinguishes safe objects from intruders. \newline \textbf{Poor:} Misidentifies or ignores key objects. \\
\textbf{6. Repetition-Token} & Identification of unnecessary redundancy in the generated reasoning. & \textbf{Excellent:} No redundancy, concise technical description. \newline \textbf{Poor:} Excessive redundancy, making reasoning unclear. \\
\textbf{7. Hallucination} & Detection of irrelevant or invented reasoning steps not aligned with visual facts. & \textbf{Excellent:} No hallucinations; grounded in the rail domain. \newline \textbf{Poor:} Majority of reasoning is hallucinated. \\
\textbf{8. Semantic Coverage} & Extent to which the response covers critical elements defined in the Ground Truth. & \textbf{Excellent:} Nearly complete semantic coverage. \newline \textbf{Poor:} Very poor semantic coverage with major gaps. \\
\textbf{9. Physics \& Momentum} & Understanding of train kinematics, 1-degree of freedom, and braking inertia. & \textbf{Excellent:} Acknowledges long braking distances and horn use. \newline \textbf{Poor:} Suggests ``Steering'' or ``Swerving'' to avoid obstacles.\\
\textbf{10. Missing Step} & Evaluation of whether any necessary logical reasoning steps are omitted. & \textbf{Excellent:} No critical steps missing. \newline \textbf{Poor:} Response is highly incomplete with critical gaps. \\
\textbf{11. Relevance} & Specificity to the scenario and correct use of railway terminology. & \textbf{Excellent:} Highly specific (e.g., uses ``Ballast'', ``Pantograph''). \newline \textbf{Poor:} Largely irrelevant or uses generic driving terms. \\
\textbf{12. Missing Details} & The extent to which critical contextual information is absent. & \textbf{Excellent:} No significant details are missing. \newline \textbf{Poor:} Response is highly lacking in necessary details. \\
\bottomrule
\end{tabular}
\end{table*}

\section*{\textbf{(II) Prompt for RailVQA-CoM Cognitive Inference}}

In the RailVQA-CoM framework, the LMM serves as the cognitive brain. The prompt fuses textual spatial-temporal logs generated by the Small Model Perception Module with visual keyframes. We carefully designed the instructions to prioritize visual evidence and enforce the structured reasoning pipeline.

\vspace{0.5em}
\noindent \textbf{System Prompt (multi-modal Fusion and CoT Enforcement):}
\begin{quote}
\itshape
You are a railway safety analysis system. Analyze visual inputs (images) and object tracking logs to provide a professional safety assessment.\\
Each image has a visual timestamp (e.g., 'Seq: 5/20') in the top-left corner.\\
\\
Input Data Instructions:\\
1. Images (PRIMARY SOURCE): The raw visual truth. Trust the images at first.\\
2. Perception Log (SECONDARY SOURCE): Generated by a weak detector, may contain errors or omissions. Use it as a reference but verify against images.\\
3. Coordinate: Coordinate Origin (0,0) is Top-Left. +X=Right, +Y=Down, corresponds to instantaneous motion of objects in logs.\\
\\
Response Format (CoT):\\
1. Perception $\rightarrow$ Reasoning $\rightarrow$ Planning $\rightarrow$ Final Answer\\
2. Final Answer: Directly answer the user's question. If it is a multiple-choice question, explicitly state the correct option.\\
Your answer must be especially concise, professional, and focused on safety implications.
\end{quote}

\vspace{0.5em}
\noindent \textbf{Defensive Prompting Mechanism:}\\
When the perception log returns ``No objects'' for more than 80\% of its entries, the system automatically injects the following critical warning into the user prompt, forcing the LMM to leverage its zero-shot anomaly detection capabilities:

\begin{quote}
\itshape
--- AUTOMATED DETECTOR STATUS: UNRELIABLE ---\\
System Report: The automated object detector found NO standard targets (People/Cars).\\
\textbf{CRITICAL WARNING}: The detector often FAILS to see non-standard obstacles (e.g., Bicycles, Rocks, Debris).\\
\textbf{INSTRUCTION}: IGNORE the detector's claim of 'No objects'. You MUST rely SOLELY on the Visual Images to find anomalies. If you see something in the image, TRUST THE IMAGE.
\end{quote}

\section*{\textbf{(III) Prompt for LLM-as-a-Judge~\cite{zheng2023judging}}}

To rigorously evaluate the reasoning capabilities, we utilized an LLM-as-a-Judge approach, which has been successfully applied in DriveLMM-o1~\cite{ishaq2025drivelmm}. The prompt defines 12 fine-grained semantic metrics, shifting the evaluation paradigm from simple n-gram matching to logic and safety-critical assessment. The evaluator was instructed with a strict system persona and the following scoring rubric (summarized in Table~\ref{tab:metrics_rubric}).

\vspace{0.5em}
\noindent \textbf{System Prompt Instructions:}
\begin{quote}
\itshape
Avoid subjective interpretation and adhere to the given thresholds. Do not add any additional explanations beyond the structured JSON output.
\end{quote}

\IEEEpubidadjcol

\bibliographystyle{IEEEtran}
\bibliography{references}

\vfill

\end{document}